# Rediscovering the Alphabet
## On the Innate Universal Grammar

*M. Yahia KAADAN and Asaad KAADAN* [1]

## Abstract

Universal Grammar (UG) theory has been one of the most important research topics in linguistics since introduced five decades ago. UG specifies the restricted set of languages learnable by human brain, and thus, many researchers believe in its biological roots. Numerous empirical studies of neurobiological and cognitive functions of the human brain, and of many natural languages, have been conducted to unveil some aspects of UG. This, however, resulted in different and sometimes contradicting theories that do not indicate a universally unique grammar. In this research, we tackle the UG problem from an entirely different perspective. We search for the Unique Universal Grammar (UUG) that facilitates communication and knowledge transfer, the sole purpose of a language. We formulate this UG and show that it is unique, intrinsic, and cosmic, rather than humanistic. Initial analysis on a widespread natural language already showed some positive results.

**Keywords**: Alphabet; natural languages; Universal Grammar; pragmatics; semantics; communication.

## Introduction

Humans have intrinsic need to answer questions and understand the universe. Through the course of human civilization, philosophers and scientists were able to answer -or at least speculate- about many fundamental questions, e.g., where did life come from? What is the age of the universe? How did it all start? And many others. One question, however, puzzled people for a long time. It is about language. They say language is the main evolutionary contribution of humans to the cosmos, and perhaps the most interesting trait that has emerged in the past 500 million years (Nowak et al. 2002). How did it all start? Was language inherited to us genetically from other species? Do we share language with other species in the first place? And how is language constructed from its basic building blocks? All these questions can be answered once we rigorously define the language.

## Language and Universal Grammar

A language is a set of sentences; A sentence is a string of symbols; And an alphabet is a finite set of symbols. Some languages are finite and some are not. There are infinitely many finite languages, as many as integers. There are also infinitely many infinite languages, as many as real numbers. Hence, the set of all languages is not countable (Nowak et al. 2002).

A grammar is a finite set of rules specifying a language. Natural languages are infinite. However, the learning theory and common sense suggest that no algorithm can learn a set of 'super-finite' languages (Nowak et al. 2002), i.e., containing all finite languages and at least an infinite one. This calls for a need to a restricted set of grammars specifying languages learnable by the human brain. The theory of this restricted set is called 'universal grammar' (UG), which was introduced by Chomsky about fifty years ago (Chomsky 1965). The remaining question is whether UG is innate or not, which is still the subject of debate. Never the less, the classical learning theory developed by Gold (Gold 1967) suggests that a restricted search space has to exist before data, i.e., the linguistic symbols, in the language acquisition process. Therefore, the term 'before data' refers to an intrinsic restricted search space, which is the innate UG. Researchers believe UG is genetically transferred from generation to generation and thus it is affected by biological evolution that results in cultural evolution of languages (Larson 2010; Lightfoot 1998). We believe, however, that UG is fixed and unique. It is not a biological phenomenon and thus does not encounter evolution. It is as fixed and unique as the mathematical logic behind it, as will be shown later. Language, on the other hand, is undoubtedly evolutionary. Genetics and environment have a profound effect on the evolution of languages.

---

[1] The authors can be reached at *y-kaadan@hotmail.com* and *asaad.kaadan@ou.edu* respectively.

## Historical Perspective

We have to distinguish language from inscription, which is no more than a linguistic expression form that may or may not succeed in delivering the message. It is notable that inscription did not express the entire capabilities of language especially at its early stages. Even though many civilizations have developed their own writing style, these styles evolved relatively differently.

One of the most important early writing styles is graphical representation, where the entire object or some parts of it were visualized graphically, e.g., drawing an entire buffalo or only its head to represent it. Abstract ideas were represented using evocative graphics, e.g., Hieroglyphs used a man with a crutch to represent *aging* and a man with his hand on his mouth to represent *hunger*.

Cuneiform writing is an advanced form of graphical representation where human voice is divided into syllables and each syllable was encoded in a graphical symbol. These symbols reached 8000 in Babylonian and Assyrian languages, each of which comprised of vowels followed by consonants. Cuneiform writing has endured a transformative evolution that we came across in 1928 in archeological excavations on the Syrian coast. In a historical site named Ugarit, many cuneiforms dating back to 1500 B.C. have been found (Day 2002; Pardee 2007). These cuneiform inscriptions consist of only 30 symbols, which is considered the first known alphabet. These inscriptions mark the astonishing moment in time where people moved from syllabic to alphabetic languages. Writing continued to evolve with time and more symbols and punctuation marks were added to accurately convey spoken language. All these consonants, vowels and punctuation marks constitute the required alphabet to accurately capture and store a language. These symbols interact with each other to convey unique ideas as accurately and rigorously as numbers interact to convey numerical values.

Similar to the fact that math consists of abstract formulas describing various physical phenomena; the language, as shown later, consists of abstract expressions describing various physical or mental elements. Words describe different states and elements, just as numbers describe different quantities. In fact, there is a great similarity between alphabets and numerical systems. Let us take the decimal numbering system for example; assuming that we did not discover the ten decimal symbols (0 to 9) and the meaning of a decimal digit, we would still be struggling with Roman numerals that describe each number with a different symbol. People will need to memorize a huge, in fact infinite, dictionary to understand different numbers. Similarly, we are struggling today with linguistic dictionaries because we do not understand the alphabet and how alphabet letters are combined to create meaningful words and sentences. Discovering what each letter means and the principles behind combining letters into language, i.e., the UG, will enable us to understand any original word or expression unless it was a man-made acronym. Language will become as simple and accurate as numerical systems.

## Universal Grammar is Unique

We introduced earlier the formal definition of a language in terms of its sub-components. The functional definition of a language, however, is a method of communication and knowledge dissemination. All living things strive for immortality. Humans, as the most advanced species, have used languages to defy time and transfer information between individuals and across generations.

Knowledge dissemination is the sole purpose of any communication process between living individuals and language is the carrier of that communication. On the cellular level, impregnation is considered a communication process where knowledge or information, DNA and RNA stored in chromosomes, are exchanged. This biological language only has a four-letter alphabet and is able to generate all complex life forms.

We believe, since language is a communication process, basic rules governing communication are the same as those governing language. These rules are what define the universal grammar (UG). Communication methods in nature range from chemical compounds in insects to air vibrations (sounds) and signals in human beings. The communication process is independent from the communication device or medium, thus the basic communication principles (what we refer to as UG) are unique. The language is independent from the means of conveying it. Based on this, an alphabet is a system that assists human beings, for example, to convey the language by sounds, graphics, signals, etc. Most researchers study languages regardless of the writing style (writing symbols), because writing or inscription is just a mean of conveying the language. Some nations even did not develop inscription symbols for their languages (Lewis 2014), because they did not have to. We will study UG regardless of the sounding symbols or phonetics as well, because sound is also a mean of conveying the language. As a summary, UG is the collection of abstract principles that govern the communication process. UG is independent from any communication device. It can be used with sounds, gestures or inscriptions. Humans, other primates or even single-cell life forms can utilize it. UG is unique. It might be inferred from our discussion that UG is life-related and thus might be biologically defined. This is not the case. In fact, we will show later that UG is mathematically defined and deeply embedded in the cosmos fabric.

# From Communication to Universal Grammar

Nouns are the most important parts of speech, and usually the naming process is un-descriptive, i.e., we use an arbitrary code or idiom to name a particular object. This is exactly what we do most of the time when we name our children, we use arbitrary names. In addition, when we transfer a word from one language to another, we use arbitrary names too, and the transferred word becomes an idiom. Nevertheless, a language with all idiomatic nouns is exhausting. Man will need to memorize a lengthy dictionary if this language was a little bit rich and comprehensive.

A better way to communicate is to give every object a descriptive name, i.e., defining that object using a short list of known, already-defined, objects; exactly as we do in numerical systems. Since the main objective of any language is to facilitate easy and accurate communication, the more easy-and-precise this language is, the closer it is to the perfect one. Based on this analysis, we consider all words in an easy-and-precise communication system to be descriptive and non-arbitrary. We will start developing our algorithm from the very basic and intuitive facts. Any rational person intuitively accepts such facts.

## Basic Definitions

One of the very basic definitions is the *element*. The element is anything you can touch, feel or imagine. Life is full of elements: human being is an element, *John* is an element, the atom is an element and the house too. Thoughts and emotions are also elements, even though their description is relative. Each element is usually comprised of parts (which are elements too.) This same element can also join other elements to form a new one. Thus, new elements are formulated using a unique process called *operation*. Operation is the second basic definition, which we call "*The Unique Creation Principle*".

Applying the principle of unique creation on two elements is called *logical conjunction* or *logical multiplication*. It is denoted by the "∧" symbol which is read as: conjugated with, or formulated with, or even "And". The "And" here represents a purely logical operation, which is the creation of a new element using two other elements. It is clear that any *operation* on different elements will usually *form* a new one. Thus, we sometimes call it a *formulation*.

The third basic definition is the *relation*, which we call "*The Identity Principle*" or "*The Unique Existence Principle*". This means the element can only exist in a unique form, i.e., it cannot exist in a form other than its own (Symbolically we write: $a = a$.) The *relation* identifies each object from another. It gives objects their "uniqueness" after being formulated by the *operation*. The *relation* never creates new elements. It only specifies existing ones. An example of this would be the "greater than" relation. This relation only defines the rank of two elements compared to each other. It does not yield a new element.

We are always in need to define new elements in order to interact with them. Each new element can be defined by figuring out its parts and the relations among them, plus the element's own relations with other elements. The new element's definition should only depend on already-defined or intuitively discovered elements. In this sense, an *element definition* is the sum of all operations that build/formulate this element from its parts, *and* the sum of all relations between this element and the other elements.

We, in order to share a knowledge with each other, e.g., sharing an address; we use a linguistic *formulation* of multiple *operations* and *relations* concerning the elements of this particular subject (or place). In short, we define that subject to others using predefined (or pre-agreed on) elements. This definition does not need to cover all aspects of the subject, only the minimum description that conveys the idea correctly. Living things are engineered to the highest efficiency in all their activities including communication. The minimum description, however, might be missing some essential facts that cause "misunderstanding" between the two parties to occur.

Based on this discussion, any information sharing/knowledge transfer act is considered a *formula*, which we call in language an *expression*. This formula or expression requires three components in order to take place: the source of knowledge/formula/expression, which we call first component; the knowledge/formula/expression itself, which we call second component; and the destination of the knowledge/formula/expression, which we call third component. The absence of any of these components will definitely cancel the communication process.

This definition implies that communication (and language) is all about the mental state of both the first and third components regarding the second one. If the speaker speaks a language unknown to the listener, then, no communication is actually taking place.

Generally speaking; The source of a (*formula*, *expression*) can be the speaker in language, the handworker in the shop, the performer on the stage, etc.

Similarly, the (*formula*, *expression*) itself could be the words (and later the inscriptions) in language, the goods in the shop, the play on the stage, etc.

And last but not least, comes the destination of a (*formula*, *expression*) which could be the listener in language, the consumer in the shop, and the critic in the audience.

## Universal Grammar is Cosmic

Obviously, each element occupies a *space*, whether physical or mental. This comes from the fact that an element can be described as *available* or *absent*. If the element is absent in some place, then it definitely exists somewhere else. The element cannot vanish and cannot come from nowhere. This is referred to in logic as "The Law of Excluded Middle" which means there is no middle (or third) state for an element; it is either existing or not.

It is also intuitive that each element has a *time* slot. This means the element cannot be available and absent at the same time. The two states must come consecutively. First, it is available and then absent or vice versa. This is referred to in logic as "The Law of No Contradiction".

The (*form, attribute*) bilateral is another substitute for the (*operation, relation*) bilateral; they both represent a general abstraction for the (*mass, energy*) concepts in physics. *Mass* is the creation of an object from its physical parts (atoms for example), whereas *energy* between these parts, or between the object and other objects, define its existing relations (atomic and other forces). To summarize, each element -physical or not- is a *formulation* of multiple operations on its parts plus the *relations* governing them. Thus, *a formula is a definition of an element and the element is a representation of that formula*. The *element* and *formula* are two faces of the same coin.

It is intuitive that each element can be described as *existing* (available) or *non-existing* (absent). The (*existence*, *absence*) states can be denoted by (1, 0) respectively. This leads us to very important concepts: *time* and *space*.

In fact, we see that *formulation* (operation) or "*Unique Creation Principle*" is another synonym for the *space* concept. In addition, the *relation* (attribute) or "*Unique Identity Principle*" is another form of the *time* concept. Without the (*formulation*, *relation*) or (*operation*, *attribute*) there will be no *space* and no *time*. This links to the interesting fact in Einstein's general relativity theory that time and space would not exist in case mass (the formulation process) vanished.

The three communication components (source, expression and destination) plus the element with its form and attributes constitute the base for our UG. They can completely and uniquely define any knowledge state and thus facilitate accurate communication, the purpose of a language. The UG elements, as shown above, are not restricted to humans, or to their intellectual merit. These elements are embedded in the universe, linked to its space-time fabric. Human beings possess UG intuitively and identify its elements as part of their existence in the universe. Similar to the fact that humans cope with basic logic and mathematics principles intuitively.

Consequently, language -just like math- is certainly discovered and not invented by humans, and the most developed language is the most discovered one. In fact, language would be available out there even if human race would not exist. This is because language, as shown earlier, is the result of mutual interaction between nature elements. Language, therefor, is *cosmic* and not *humanistic*. Since many other creatures, especially high primates like Chimpanzees, have shown basic logic inference capabilities, we believe they are also equipped with the same UG. This means we can, in theory, communicate with them given suitable devices and deeper understanding of their alphabet. The only things separating us from other primates in terms of language are the following:

- We use a different "coding" scheme, i.e., we encode our alphabet with different sounds, gestures, etc.
- We have much more processing capabilities and thus we can further extrapolate ideas and combine abstract thoughts to reach deeper understanding of our environment and communicate this understanding more accurately. Thus, in this work, we link language development directly with brain size and capabilities.

It is worth mentioning that quantum physics breaks many of the classical physical laws. As a result, the above laws of Excluded Middle and No Contradiction do not exist in the world of small particles. Quantum phenomena, however, never appear in our macroscopic world. Understanding the bizarre quantum phenomena is still a work in progress and it is not known if this behavior is actually happening or it is just an illusion created by our measurement devices working at the quantum scale. Since language is used to describe our macroscopic world, quantum physics do not pose any threats for UG.

# Analyzing the Universal Grammar

Let us study the formula of an element **F** from a set **U**. We represent existence (or availability) of element **F** by drawing some boundary inside set **U**. As seen in Figure 1.

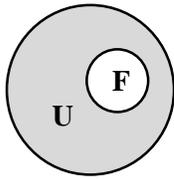

**Figure 1**

Thus, the element's absence is represented by **F** *complement* with respect to **U**. This is the shaded area between the two boundaries. Since **F** is absent from the shaded area (**F** complement), then it is definitely available in the non-shaded area (**F**) according to the "The Law of Excluded Middle".

It is clear that when we look at the area (or space) associated with element **F** we determine its form, i.e., the elements that formulate **F**. However, when we look at the space associated with **F** complement, we determine the form of elements inside **F** complement, i.e., outside **F**. Determining the form of the complement's elements leads to the identification of any element outside the complement, i.e., inside **F**. The *identification* of an element does not provide us with its specific form rather with its attributes that help identify it from others. Based on this logic, an element is determined (i.e., its form) using its own space. An element is identified (i.e., its attributes) using its complementary space. Intuitively, **F** takes a value '1' inside its own space and a value '0' inside its complementary space.

One important aspect to mention is that determining an element's form using its space alone does not identify it. In order to identify this element we need to use its complement space too. The complement represents a comparison system used in the identification process. For example, we cannot identify "Beauty" if we do not know "Ugliness". Ugliness is the comparison system that gives beauty its meaning. On the other side, identifying an element using its complementary subspace alone does not determine that element for us. If you knew "ugliness" alone without determining any form of "beauty", then beauty is not known to you.

The previous discussion leads us to the following fact: Our knowledge about an element can be completely and accurately classified into four different states: Knowing only the form, knowing only the attributes (relations), knowing both the form and attributes, or knowing none of them. Mathematically, it is the knowledge of **F**, **F** complement, both **F** and its complement, or none of them, respectively.

We usually denote elements with small English letters, e.g., a, b, c, etc. Each element according to the Excluded Middle Law must be in either one of the existence/absence states, i.e., 1 or 0, respectively. Let us take the element a. The complementary element is denoted "~a" and according to Excluded Middle Law they cannot coexist in the same place, i.e., If (a) is available here then ~a is not. The element (~a) is definitely available in some other place (It cannot vanish.) Note that according to The Law of No Contradiction the element must move to the opposite state if the values were to be changed.

As a summary, the following basic definitions are used as a base for our UG:

- The *element* or *formula*.
- The *operation*, *form*, or the *Unique Creation* principle.
- The *relation*, *attribute*, or the *Unique Identity* principle.

Our knowledge about an element, i.e., its form and attributes, is governed by two fundamental concepts: the Law of Excluded Middle (the *space* concept) and The Law of No Contradiction (the *time* concept).

## Combining Formulas

Since formulas (elements) constitute our basic building blocks, we will introduce in this section the *formula composition* principle, which will be of a great importance in developing our formal UG theory in the following sections.

Let us take the formulas $f_1$ and $f_2$. We can create a new formula via the composition of these two formulas. The following examples explain the composition process in detail:

If (x $f_1$ y) was (x is y's brother) and (y $f_2$ z) was (y is z's mother),
then (x is z's uncle) is (x $f$ z), where $f = f_1 \circ f_2$; We read "$f_1$ composition $f_2$".

If (x is y's mother) was (x $f_2$ y) and (y is z's brother) was (y $f_1$ z),
then (x is z's mother) is (x $f$ z), where $f = f_2 \circ f_1$; We read "$f_2$ composition $f_1$".

The previous examples show that in general $f_1 \circ f_2 \neq f_2 \circ f_1$. We can apply the composition principle to more than two formulas. For example: $f = f_1 \circ f_2 \circ f_3$. This brings us to the *associativity* property of composition: $(f_1 \circ f_2) \circ f_3 = f_1 \circ (f_2 \circ f_3) = f_1 \circ f_2 \circ f_3$. Thus, we can drop all parentheses without affecting the results.

We can also do a composition of a formula with itself, i.e.:

$$f_1 \circ f_2 \circ \ldots \circ f_r \circ \ldots \circ f_n = f^n, \text{ when } f_1 = f_2 = \cdots = f_r = \cdots = f_n.$$

For example:

If (x is y's father) was (x $f$ y), then (x is z's grandfather) is (x $f^2$ z).

In formula composition, destination (output) of the first formula becomes source (input) of the second one. Thus we write $f_2(f_1(x))$ to describe $(f_1 \circ f_2)(x)$, i.e., we first look for $f_1(x)$ and then for $f_2(f_1(x))$. Also, we write $f_1(f_2(x))$ to describe $(f_2 \circ f_1)(x)$, i.e., we first look for $f_2(x)$ and then for $f_1(f_2(x))$. In general, $f_1 \circ f_2 \neq f_2 \circ f_1$. We can also write without parentheses $f_1 \circ f_2 = f_2 f_1$, $f_2 \circ f_1 = f_1 f_2$.

## General Structure of Linguistic Expressions

We are now ready to study the general structure of any linguistic expression. Language is used by the speaker to convey or transfer a specific subject (or idea) to the listener. This transfer can be achieved by sounds, gestures, etc. These are all considered language.

<u>Postulate</u>: Language is a set of formulas.

We say language is a set of formulas and not the set of *all* formulas; because we can proof mathematically there is no such set of all sets (or all formulas). No matter how many formulas are included in one set, we can always find other formulas that are not included in it (Notice the agreement with our earlier language definition.)

Since language is used to transfer knowledge, i.e., formulate that knowledge in a form completely understood by the listener, we can describe it using formulation principle. If we write a formula $f$ using its components $f_1, f_2, \ldots, f_n$ then we get the general expression for the formulation principle, i.e., the general expression for language:

$$y = f_n\left(\ldots\left(f_2(f_1(x))\right)\right).$$

As we showed earlier, formula composition is an associative operation, thus we can write the previous expression without parenthesis:

$$y = f_n \ldots f_2 f_1(x).$$

<u>Corollary 1</u>: The general structure of any linguistic expression is $f_n \ldots f_2 f_1$, i.e., a composition of basic formulas $f_1, f_2, \ldots, f_n$. We call these basic formulas *alphabet letters*. Note the similarity with numbers. Each number, e.g., 3657, is formulated by composing *digits* or basic symbols in this numbering system with each other.

## Communication States and the Alphabet

The formula that defines an element is relative to both the speaker and the listener, i.e., each one (or both) of them may identify only the form, or only the attributes, or both form and attributes of that element. They might also misidentify both the form and attributes of the element.

In addition, when we define an element we tend to identify its components, the relations between these components, and the element's relations with other elements. This definition is relative because there is unlimited number of components for each element and unlimited number of relations among them. We usually, due to instinctive power-efficient behavior, use the minimum acceptable description for each element. This minimum description, however, should be completely understood by the listener, i.e., based on a common knowledge between the communication parties.

| a | b | c |
|---|---|---|
| 1 | 1 | 1 |
| 1 | 1 | 0 |
| 1 | 0 | 1 |
| 1 | 0 | 0 |
| 0 | 1 | 1 |
| 0 | 1 | 0 |
| 0 | 0 | 1 |
| 0 | 0 | 0 |

**Table 1  Three-element Truth Table**

Let us use the letters **a, b, c** to represent a speaker (source), a subject, and a listener (destination) respectively. We can encode the presence and absence of these three elements in eight distinctive states, described in the truth table (Table 1). Furthermore, the subject (or element) can appear in four distinctive states for both the speaker and the listener: either a form, or an attribute, or both a form and an attribute, or none of them. This results in four permutations for each line in Table 1. This raises the total number of distinctive communication states to 32 (Table 2). Each one of these states describes a unique communication state between the speaker and the listener regarding the subject of interest. We denote the form (formulation) and attribute (relation) with $*$ and $\mathfrak{R}$ respectively.

Logical development in the previous sections led us to identify the principal components of any communication process: The first element (source), the second element (subject), and the third element (destination). Communication about a subject requires defining its form and attributes as seen or understood by both the source and destination. If we assign to each one of these components a "1" for existence and a "0" for absence, then we will get a five-bit binary code describing each and all-possible communication states as shown in Table 2. Theses bits from left to right represent the source, the subject, the destination, form of the subject, and attributes of that subject, respectively.

| a | b | c | * | ℜ | # | a | b | c | * | ℜ | # |
|---|---|---|---|---|---|---|---|---|---|---|---|
| 1 | 1 | 1 | 1 | 1 | 1 | 0 | 1 | 1 | 1 | 1 | 17 |
| 1 | 1 | 1 | 1 | 0 | 2 | 0 | 1 | 1 | 1 | 0 | 18 |
| 1 | 1 | 1 | 0 | 1 | 3 | 0 | 1 | 1 | 0 | 1 | 19 |
| 1 | 1 | 1 | 0 | 0 | 4 | 0 | 1 | 1 | 0 | 0 | 20 |
| 1 | 1 | 0 | 1 | 1 | 5 | 0 | 1 | 0 | 1 | 1 | 21 |
| 1 | 1 | 0 | 1 | 0 | 6 | 0 | 1 | 0 | 1 | 0 | 22 |
| 1 | 1 | 0 | 0 | 1 | 7 | 0 | 1 | 0 | 0 | 1 | 23 |
| 1 | 1 | 0 | 0 | 0 | 8 | 0 | 1 | 0 | 0 | 0 | 24 |
| 1 | 0 | 1 | 1 | 1 | 9 | 0 | 0 | 1 | 1 | 1 | 25 |
| 1 | 0 | 1 | 1 | 0 | 10 | 0 | 0 | 1 | 1 | 0 | 26 |
| 1 | 0 | 1 | 0 | 1 | 11 | 0 | 0 | 1 | 0 | 1 | 27 |
| 1 | 0 | 1 | 0 | 0 | 12 | 0 | 0 | 1 | 0 | 0 | 28 |
| 1 | 0 | 0 | 1 | 1 | 13 | 0 | 0 | 0 | 1 | 1 | 29 |
| 1 | 0 | 0 | 1 | 0 | 14 | 0 | 0 | 0 | 1 | 0 | 30 |
| 1 | 0 | 0 | 0 | 1 | 15 | 0 | 0 | 0 | 0 | 1 | 31 |
| 1 | 0 | 0 | 0 | 0 | 16 | 0 | 0 | 0 | 0 | 0 | 32 |

**Table 2 The Communication States**

Let us take an example the code "10110". We start with the leftmost bit. It represents the availability (existence) of source "1"; the absence of subject "0"; the availability of destination "1"; the mutual availability "1" of form of the absent subject in both minds of source and destination; and finally the rightmost bit represents the mutual absence "0" of attributes of the absent subject in both source and destination minds. Note that form and attributes of a subject do not appear or disappear on their own. These are relative concepts described only with regard to the mental state of both source and destination. As a result, language does not only care about the speaker mental state but also the listener.

Now, if we give each distinctive communication state a symbol that identifies it completely from other states, we get the complete set of communication alphabet. This alphabet is based on the intrinsic and innate principles of UG introduced above, thus, it is capable of transferring knowledge completely and accurately. Analyzing the abstract meaning of the code "10110" above seems difficult and un-natural. Not to mention, linking this abstract concept to our physical world and combining tens or hundreds of these codes at a time. Language is, indeed, a great feat. However, our human brain, and maybe other advanced brains, is so complex and so powerful. It is customized to analyze these abstract codes and extract their meanings on the fly. Remember that many of other human, and non-human, activities are actually extremely complex, e.g., moving your body from point A to point B involves complex kinematics and dynamics calculations to control hundreds of muscles; visual tracking involves storing, processing and analysis of huge amounts of visual data, etc. We perform all these activities without breaking a sweat because our brains are designed to do so, and language, is not, by any means, different. The human brain, and probably other brains, come already equipped with the language engine that can decode this cosmic UG. This engine interfaces with another encoding/decoding modules that specify how to physically convey this language using sounds, gestures, inscriptions, etc.

Figure 2 shows a suggested block diagram for the language process inside our brains. The language decoders perform the task of converting physical signals into alphabet letters or digital codes. These concatenated codes represent composed formulas as we showed earlier. These formulas are fed into the UG engine that convolutely parses each one of them until a single or multiple values are calculated for the entire expression. These values are then analyzed in order to take actions, such as retrieving a memory, performing numerical calculation, logical inference, etc. The results of these actions need now to be formulated in a suitable expression containing all needed information. The encoders encapsulate the expression digital codes with physical signals to be transmitted by the appropriate device. The contents of the UG engine are speculative and not fully understood. Both modules that parse or build an expression depend on general principles of UG for their operation. Note that all modules in Figure 2 are biological except the UG. UG is a set of fixed, universal, cosmic rules that define how language is parsed on the most abstract level. All other modules can evolve and differ between creatures, generations and even individuals. This structure explains why infants (Kennison 2014; Sakai 2005) can learn any language better than adults. The only logical explanation is that their language encoders/decoders are not defined yet. These modules are defined by the environment and they can be easily replaced at an early age. Once the environment defines these modules, repetition and parents/teachers help solidify these encoders/decoders in the infant neural structure so that it becomes difficult to change them. What about the UG engine itself? It was transferred to us through our genes and this is the biological root that current researchers believe UG might have.

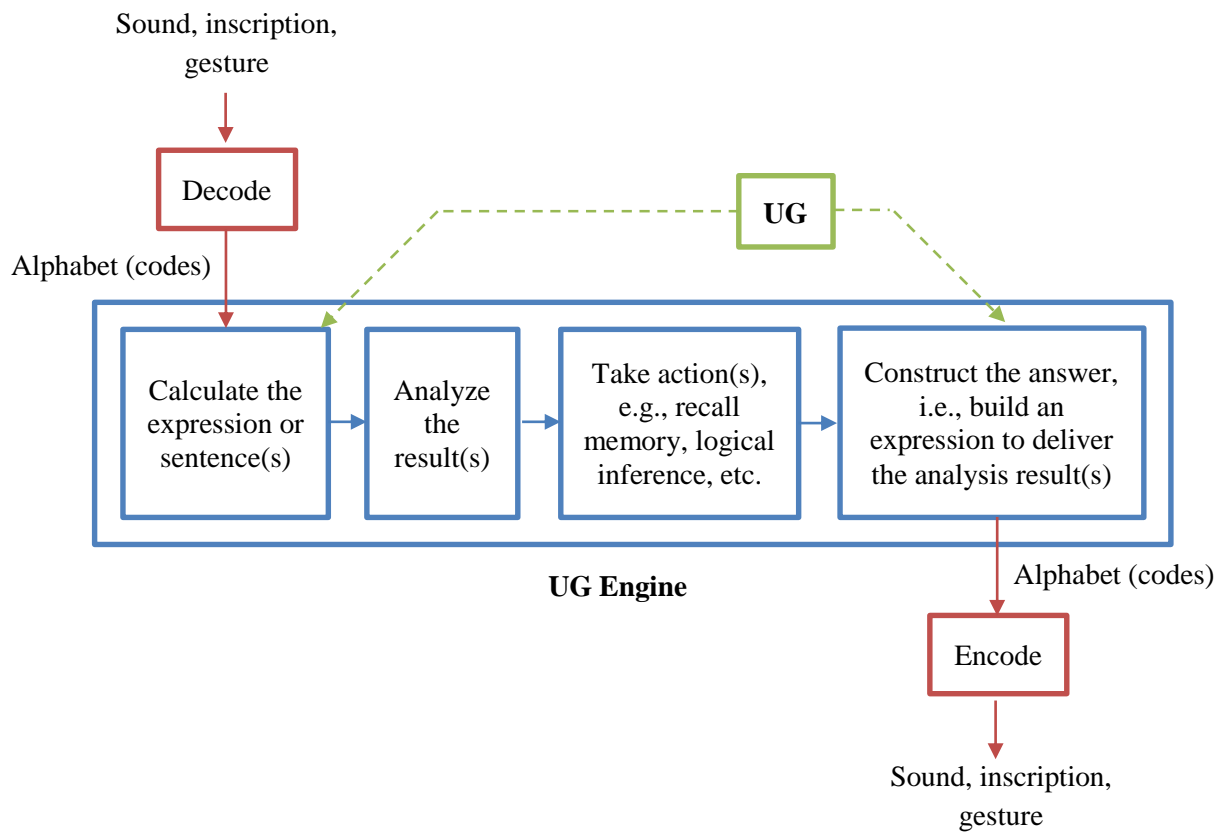

**Figure 2  Language Process Block Diagram**

# UG-compatible Natural Languages

Based on the discussion above, we can, in theory, build a formal language that embodies all UG rules and thus achieves the perfect communication. The important question would be; what about natural languages? Do UG-compatible natural languages actually exist? This is the same decades-old question in linguistics about whether formal natural languages exist or not. Before we present our results, it is worth to say that whether there are truly formal natural languages or not, we believe some or many natural languages should be at least near-formal, given that nature always strives for optimality. A formal natural language is the optimal solution for communication and hundreds of thousands of years of homo-sapiens evolution should result in a near-optimal communication solution.

The 32 digital codes in Table 2, or the communication alphabet, constitute what we call the *Digital Alphabet*. It is intriguing to check whether this digital alphabet maps partially or totally to any of the existing natural alphabets. Since our first language is Arabic, which is one of the old spoken Semitic languages (Woodard 2008), it made sense to start with it as a good candidate. The results were, indeed, interesting; however, not conclusive. We are conducting extensive research to understand more about UG and the digital alphabet. It is worth to mention that many of our results may apply to other languages as well, especially the old ones that have not been affected much by man-made artificial grammar and synonyms. It will be just a matter of figuring out the encoding of this language, i.e., how its alphabet is mapped to the digital alphabet. In our case, it was straightforward and required only a couple of iterations. Again, we should not expect natural languages to be fully formal or UG-compatible. There is always a place for randomness in nature.

Combining Table 2 with the Arabic alphabet ordered using the *abjadi* order yields Table 3. Rearranging Table 3 in eight rows and four columns yields Table 4, which is used throughout the paper. It describes $2^5 = 32$ unique communication states and their appropriate Arabic letters. Surprisingly, Arabic alphabet has 28 consonants + 4 vowels (three vowels and the absence of all vowels) = 32 unique symbols. These 32 different codes will be used to generate a formal structure to build names and language in general. We ordered the *Digital Alphabet* from **11111** to **00000** based on mathematical development of truth tables, while the Arabic letters were sequenced according to *abjadi* order, which is the historical sequence of Arabic alphabet and most ancient alphabets derived from the Ugaritian one (Daniels 2013).

| a | b | c | * | ℜ |   | # | a | b | c | * | ℜ |   | # |
|---|---|---|---|---|---|---|---|---|---|---|---|---|---|
| 1 | 1 | 1 | 1 | 1 | ا | 1 | 0 | 1 | 1 | 1 | 1 | ف | 17 |
| 1 | 1 | 1 | 1 | 0 | ب | 2 | 0 | 1 | 1 | 1 | 0 | ص | 18 |
| 1 | 1 | 1 | 0 | 1 | ج | 3 | 0 | 1 | 1 | 0 | 1 | ق | 19 |
| 1 | 1 | 1 | 0 | 0 | د | 4 | 0 | 1 | 1 | 0 | 0 | ر | 20 |
| 1 | 1 | 0 | 1 | 1 | هـ | 5 | 0 | 1 | 0 | 1 | 1 | ش | 21 |
| 1 | 1 | 0 | 1 | 0 | و | 6 | 0 | 1 | 0 | 1 | 0 | ت | 22 |
| 1 | 1 | 0 | 0 | 1 | ز | 7 | 0 | 1 | 0 | 0 | 1 | ث | 23 |
| 1 | 1 | 0 | 0 | 0 | ح | 8 | 0 | 1 | 0 | 0 | 0 | خ | 24 |
| 1 | 0 | 1 | 1 | 1 | ط | 9 | 0 | 0 | 1 | 1 | 1 | ذ | 25 |
| 1 | 0 | 1 | 1 | 0 | ي | 10 | 0 | 0 | 1 | 1 | 0 | ض | 26 |
| 1 | 0 | 1 | 0 | 1 | ك | 11 | 0 | 0 | 1 | 0 | 1 | ظ | 27 |
| 1 | 0 | 1 | 0 | 0 | ل | 12 | 0 | 0 | 1 | 0 | 0 | غ | 28 |
| 1 | 0 | 0 | 1 | 1 | م | 13 | 0 | 0 | 0 | 1 | 1 | ◌ُ | 29 |
| 1 | 0 | 0 | 1 | 0 | ن | 14 | 0 | 0 | 0 | 1 | 0 | ◌َ | 30 |
| 1 | 0 | 0 | 0 | 1 | س | 15 | 0 | 0 | 0 | 0 | 1 | ◌ِ | 31 |
| 1 | 0 | 0 | 0 | 0 | ع | 16 | 0 | 0 | 0 | 0 | 0 | ◌ْ | 32 |

**Table 3  The Digital Alphabet**

The last row in Table 4 describes the vowels (diacritics) or *ḥarakāt* (حَرَكَاْتْ) in Arabic. They are sequenced as following: *ḍammah*, *fatḥah*, *kasrah* and *sukūn*. Table 4 is also called the *Alphabetical Truth Table*. Since Arabic alphabet maps completely to the digital alphabet and thus encodes all possible communication states between a speaker and a listener, *it completely covers the knowledge theory*. Our theory proves that Arabic (at least the core unmodified part of it) is indeed a semantic language and not a syntactic one. It is a formulation of logical components and not a combination of phonetics.

The *Digital Alphabet* starts with **11111**, which we name *'alif* ( ا ). It is derived from the Arabic word *'elfah* (إِلْفة) or *affinity*. It means all three formulation elements are present (the first three ones from the left), and both source and destination have a complete mutual knowledge about the subject including its form and attributes (the last two ones). In other words, both source and destination have a complete agreement or *affinity* about this subject. The digital alphabet ends with **00000** or *sukūn* (سُكُوْنْ). It means silence or more precisely *nullity*. There is simply nothing, everything is absent including the speaker.

| dāl د | jīm ج | bā' ب | 'alif ا |
|---|---|---|---|
| **11100** | **11101** | **11110** | **11111** |
| ḥā' ح | zayn ز | wāw و | hā' ه |
| **11000** | **11001** | **11010** | **11011** |
| lām ل | kāf ك | yā' ي | ṭā' ط |
| **10100** | **10101** | **10110** | **10111** |
| 'ayn ع | sīn س | nūn ن | mīm م |
| **10000** | **10001** | **10010** | **10011** |
| rā' ر | qāf ق | ṣād ص | fā' ف |
| **01100** | **01101** | **01110** | **01111** |
| khā' خ | thā' ث | tā' ت | shīn ش |
| **01000** | **01001** | **01010** | **01011** |
| ghayn غ | ẓā' ظ | ḍād ض | dhāl ذ |
| **00100** | **00101** | **00110** | **00111** |
| sukūn ْ | kasrah ِ | fatḥah َ | ḍammah ُ |
| **00000** | **00001** | **00010** | **00011** |

**Table 4 The Digital Alphabet Table Rearranged**

Using Table 4, we find that *fatḥah* (**000**10) is nothing but a formulation/operation (form), *kasrah* (**000**01) is a relation (attribute), *ḍammah* (**000**11) is both a form and attribute, and *sukūn* (**000**00) is neither form nor attribute, it is nothing. Since the element itself and the two communicating parties are absent from these four letters, we find the diacritics to define the element's space/time state. The diacritics in space describe the availability or absence of form and attributes; while in time, they describe the sequence of appearance of both form and attributes.

*Fatḥah* denotes *past* because it describes a form that already existed. Indeed, all past tense verbs in Arabic must end in *fatḥah*. Ending the word with *fatḥah* means that we started formulation with it, since formula composition works from the last written (inner) formula to the first written (outer) one as shown ealier. This means the speaker started this formula (word) by telling the listener that it already happened in the past. *Kasrah*, on the other hand, denotes *future* because it describes the attributes of an object only, i.e., it identifies the object before it takes a form, which is going to happen in future. The single-letter imperative verbs in Arabic always end with *kasrah* in order to make clear the imperative future tense of the verb, e.g., *ri* (ر) or *look* and *qi* (ق) or *protect*. *Ḍammah* denotes *present* because it describes both form and attributes of an object, i.e., it is past and future occurring together and that is the present tense. All present tense verbs in Arabic must end in *ḍammah* as well. Finally, *sukūn* denotes time and space independence. It is the absence of form and attributes together at the same time, which leads to a timeless/placeless definition. This is usually the "abstract" definition of an element. In fact, all verbal nouns in Arabic (the time independent versions of these verbs) end in *sukūn*.

Given that diacritics describe the element (formula) space/time definition, we should formulate any formula using diacritics, i.e., they should be applied to all formulas (expressions) because all formulas satisfy the space and time principles. Linguistically, we find that any Arabic word or expression has diacritics, whether they were written or not.

The short discussion above illustrates the powerful results that can emerge from deep analyses of the UG and the digital alphabet. There are much more to discover about language, grammar and how we process and analyze our environment. The next section presents many more results and examples that help understand more about UG and the digital alphabet, and at the same time, explain many of the Arabic language properties that have been mysterious to researchers for hundreds of years.

## Results and Examples

We will study in this section some single-, double- and triple-letter Arabic formulas as a tutorial on how to research and analyze a language using our developed UG and the digital alphabet. There will be many Arabic words throughout the discussion. The English translation is provided in *italics* wherever possible. Readers unfamiliar with Arabic language, however, can simply skip this part. It is only intended to present a few examples out of countless others.

Before we start we need to reemphasize an important concept; Formulation, as we said earlier, starts with the last formula and ends with the first one: $f_1 \circ f_2 = f_2 f_1$, $f_2 \circ f_1 = f_1 f_2$. This means if we want to construct a word (formula) from the two letters $f_1$ and $f_2$ (formulas too), we have to write $f_2 f_1(x)$, i.e., the second letter $f_2$ is written first and then the first letter $f_1$. It also applies to Arabic, formulation starts with the last written letter and ends with the first written one.

Every sentence is a formula comprised of multiple words (formulas) and each one of them is comprised of multiple letters, which represent the basic (atomic) formulas.

We found the letter *'alif* (**11111** ا) to give the affinity meaning, i.e., the complete knowledge and agreement of both communication parties about the available communicated subject. The letter *bā'* (**1111**0 ب) is the knowledge of both communication ends of only the form of an available subject. Letter *jīm* (**111**0**1** ج) on the other hand, is the knowledge of both communication ends of only the attributes of an available subject. The element's formulation is what gives its shape, template, or definition. When the element itself is available, then describing its formulation without any specific attributes, as in letter *bā'*, means the element is undergoing transformation; or it had already transformed into its current form but we cannot identify its specific details (attributes) yet. This is why the letter *bā'* appears at the beginning of many words involving transformation from one form into another, e.g., *ended up* (باء), *sold* (باع), *cried* (بكى), *moved far away* (بَعُد), etc. In the case of *jīm*, we cannot identify the available subject form because it is either too big, or too wide, or too far, etc., to be comprehended by humans. If the subject is not physical, maybe it cannot take a specific form. Examples include words such as *paradise* (جنة), *hell* (جهنم/جحيم), *demons* (جن), *mountain* (جبل), etc.

Both letters *bā'* and *jīm* can be interpreted in a different way based on our previous space/time discussion. The form transformation with *bā'* refers to a spatial/physical transformation, i.e., the element is moving into or out of this space/form regardless of time. The emphasize, here, is not about the *change*, it is about the *result*, which is general and not completely identified. In the case of *jīm*, where attributes transformation refers to a change in time, the element is moving into or out of this time slot regardless of its space/form. The emphasize, here, is not about the *result*, it is about the *change*. These concepts show up clearly in the following examples. The word *ended up* (باء) refers to a change in form or state where the surrounding words usually specify the new form/state, i.e., the transformation result. The word *came* (جاء) on the other hand, refers to a change of location with time, i.e., mobility, and surrounding words usually specify the new/old place to emphasize the transformation itself.

There is no problem about different interpretations of the same formula as long as they do not logically contradict. This is a norm in math since it is an abstract description of our physical world. A second-order equation may be interpreted as a projectile trajectory, a current/voltage relationship in a circuit, or describing change of temperature in a living room. We cannot say, however, that it describes a straight line because this statement contradicts the equation's basic abstract second-order properties.

The combination of these simple 32 letters in a formula opens the door for endless interpretations and possibilities. Some formulas are already used and widespread, some of them are seldom used and others no body heard about them at all. Take for example the word *door* (باب). If we take the *space* interpretation of letter *bā'*, then the formula refers to an affinity (*'alif*) between two spaces (the two *bā'* letters). A door allows the communicating parties to share complete information/knowledge/value between two places, two worlds, or two different forms of things. This is the most abstract and accurate definition of doors that one can come up with. If we experiment a little and replace the two *bā'* letters with *jīm* letters, we expect to get a formula that connects two elements in two different time episodes. This formula is (جاج). It is not used in Arabic. However, small modifications will turn this unused formula into a widely known word. Add the letter *'ayn* (ع **100**00) for example to get *traces* (عجاج). It is a word used to describe dirt and sand coming from the ground when there is something large moving (an army, a caravan, a horse, etc.) Notice that this movement described by (جاج) is only available to the speaker and the moving subject itself is not available. This word is only used for hidden subjects. You see the traces of the army moving but the sand and dirt cloud is so large (meaning a lot of momentum and power) that it is hiding the army itself. If you use the letter *ḥā'* (ح **110**00), the only difference is that the moving element is actually available with the speaker only and not the listener. This means the element *came* to the speaker and this *coming* involved mobility (change with time). The word (حجّاج) means *pilgrim* in Arabic, which is exactly what the formula described. (The extra *jīm* is used to emphasize mobility.)

One of the important statements that one can make about UG is that it describes everything was, is, and going to be said in language. It allows us to make new words for future needs just as easy as we make larger and larger numbers whenever we need them.

There are many words (formulas) starting with *'alif*, or using traditional definitions ending with letter *'alif*. Every formula starting with *'alif* (a word ending with *'alif*) means both speaker and listener have affinity with the second (available) element or the subject which this word is describing. The subject could be a singular masculine one, e.g.: *threw* (رمى : رما), *gave* (أعطى : أعطا), *boy* (فتى : فتا), etc. It can also be a singular feminine, e.g.: *Layla* (a name) (ليلى : ليلا), *thirsty* (عطشى : عطشا), etc. It is possible to have a dual or plural subject, e.g., *(they both) hit* (ضربا), *(they both) wandered* (سعيا), *hills* (رُبى : رُبا), etc. These examples show that starting a formula with *'alif* does not have any duality, plurality, or feminine effect (as usually believed in Arabic grammar). It is used to show the affinity of both the speaker and listener with the subject described by this word. This affinity takes place with an available subject, which its form and attributes are completely known by both communication ends.

There are many formulas starting with two *'alif* letters as well, or as traditionally said, words with *'alif* and a *hamzah* at the end (*hamzah* is an *'alif* with *sukūn*, which adds an extra meaning but we will not go here into more levels of details to save time.) The first *'alif* is an affinity of both the speaker and listener with the available subject. The second *'alif*, which is formulated/combined with the first one, is also an affinity of both the speaker and listener with an available subject. This subject itself is an affinity with another available subject given the first *'alif*, i.e., it is an affinity combined with another affinity. This results in widening the effective space of this affinity, i.e. targeting a wider space. The following examples clarify this point: (*came* جاء, *desert* صحراء, *sky* سماء, *enemies* أعداء, *water* ماء, etc.) It is obvious that the described subjects are either wide, large, far or simply uncountable/unmeasurable. Thus, the speaker had to use two *ʾalif* letters, or double affinity, to make sure the listener understands the described subject clearly (and understands its dimensions).

After we studied the effect of *ʾalif* at the beginning of the formula, let us study its effect at the end of it. Every formula ending with an *ʾalif* (a word staring with an *ʾalif*), describes a mutual definition, i.e., a mutual understanding between the speaker and listener about the entire form and attributes definition of an available subject, which is the formula itself. In other words, adding *ʾalif* at the end of the formula (at the beginning of the word) implies that both communication ends now have a mutual comprehensive definition of that formula or word. The following examples show how the formula is shared between the two communication parties by introducing an *ʾalif* to the end of it: (*preparation* إعداد, *came* أقبل, *hit* إضرب, *dedication* إهداء, announcement إبلاغ, etc.)

As a result, we see that ending a formula with *ʾalif* is a sign of mutual definition or mutual understanding between the speaker and listener about the available subject (the formula itself). It is not a sign of a singular, a feminine, or a plural subject. Here, also, we find many words that start with two *ʾalif* letters, or formulas ending with two *ʾalif* letters. Using a similar analysis, we conclude that starting a word with two *ʾalif* letters is nothing but a mutual definition of a mutual definition of that word, i.e., expanding the space of mutual definition and emphasizing it. For example: *later* (آجل : أأجل), *ruins* (آثار : أأثار), *sinner* (آثم : أأثم), *coming* (آتي : أأتي), etc.

The use of letter *ʾalif* is not restricted only to the beginning and end of the word, rather inside it too. The meaning will change depending on *ʾalif* position and on the letters before and after.

Let us take the following formula *two men* (رجلان). The letter *nūn* (ن 10010) means that only the speaker (because the listener is absent) knows only the form of an absent element. Knowledge of element's form only, means knowing the general template of that element, i.e., the speaker knows this element generally, without details. An explicit example for this is the formula *a man* (رجلٌ = رجلن). Here we started the formula with letter *nūn* to imply that the formula (رجل) is general and not specific (indefinite form). It is only a form of the formula *man* (رجل), i.e., the general or template case without specifying a unique man.

For any countable formula to be general (indefinite), it should describe at least two elements. If it describes only one, then the formula is unique and identified. If we combine the letter *nūn* with *ʾalif* as in (ان), then the subject which was previously absent and known only generally with *nūn*, is now available and completely known to the speaker and listener with *ʾalif*. The introduction of affinity to the general knowledge means this formula describes only two elements, not less and not more, in case the element was countable as in *two men* (رجلان). Describing a single element contradicts with the general knowledge brought by *nūn* and describing more than two elements does not allow for a complete affinity as expressed by *'alif*.

In case the subject was uncountable, the formula describes an unspecified widely known element that is the subject itself, as in *full (with food)* (شبعان). The subject here (شبع) is a general formula but completely known to both speaker and listener. The following examples show the effect of (ان) in countable and uncountable subjects: (فان *mortal*, إثنان *two*, رجلان *two men*, شبعان *full*, أشجان *sorrows*, هاتان *these two*, etc.)

To summarize, the use of (ان) at the beginning of a formula denotes the definition of a subject which was previously absent and known generally by the speaker via its form and now it is available and completely known by both communication parties. This abstract definition of (ان) does not specify whether the subject is singular or plural, whether it is dual or more, etc. The nature of the formula itself determines such details.

Moving forward, let us take the formula *the book* (الكتاب). We study it as follows: ((ا) ل (كتاب)) where we consider the formula *book* (كتاب) to be already known and defined. As usual, we start with the beginning of the formula (the end of the word). The letter *lām* (**101**00 ل) means the previous formula *book* (كتاب) is nothing but an unavailable subject that the speaker and listener know neither its form nor its attributes, i.e. the formula *book* (كتاب) is neither known nor identified by communicating parties. Now we add the letter *'alif* and the previously absent and unknown formula *book* (كتاب) becomes available and completely known to both of them. We conclude that formulating any formula with *the* (ال) means the affinity of communication parties with an available formula that was previously absent and unknown to both of them, i.e., the tool *the* (ال) defines and specifies this formula completely; hence, we call it the *definition tool*.

Now let us take this formula *do not play* (لا تلعب). We study it as follows: ((ل ( ا )تلعب)) where we consider the formula *play* (تلعب) to be already known and defined. Starting with *'alif*, both the speaker and listener have a complete affinity and knowledge of the previous formula *play* (تلعب). Adding the letter *lām* (**101**00 ل), the known formula becomes absent and completely unidentified by communication parties. This means the tool *not* (لا) took a previously available and known formula and transferred it to an unavailable and completely unknown one, i.e., the tool *not* (لا) negated or canceled the definition of that formula; hence, we call it the *negation tool*. Note that formulation with *the* (ال) is exactly contrary to the formulation with *not* (أْلَ = لا).

We will start now analyzing some bi-letter formulas where we fix the second letter (*lām*) and change the first letter to study the effect on meaning. We start with the formulas (*the* أَلْ, *showed up* هَلْ, *walked away* فَلْ, *appeared* طَلْ). We studied the first formula *the* (أَلْ) earlier and found its meaning to be the complete affinity of both speaker and listener with an available subject that was previously unavailable and unknown for both of them, i.e., they both defined the subject clearly. In the second formula *showed up* (هَلْ), the letter *hā'* (**11011** هـ) means only the speaker knows the complete form and attributes of an available subject. This means the speaker here attained a complete knowledge about an available subject that was previously unavailable and unknown for both communication parties, i.e., the subject came close to the speaker physically or mentally.

Now moving to the formula *walked away* (فَلْ), and looking at the letters' *fā'* (**01111** ف) and *lām* (**101**00 ل); We see that only the listener here gained a complete knowledge about an available subject that was previously unavailable and unknown for both communication parties, i.e., the subject escaped/sneaked toward the listener physically or mentally. In other words, it *walked away* from the speaker.

It should be clear now the movement introduced in the three formulas (*the* أَلْ, *showed up* هَلْ, *walked away* فَلْ) by exchanging the letters *'alif*, *hā'* and *fā'*. First the subject was absent using *lām*, and then it became available and known by either both the speaker and listener (*'alif*), or the speaker alone (*hā'*), or the listener alone (*fā'*). The movement here is short and enough to transfer the subject from one state to another. Another set of letters cause a large movement that gives different physical and mental meanings.

Now let us take the fourth formula *appeared* (طَلْ). Using the letter *ṭā'* (**10111** ط) definition, we find its meaning to be the speaker and listener knowledge of the form and attributes of an unavailable subject that was previously unavailable and unknown for both of them. This means the speaker and listener recognized that subject without seeing it closely, i.e., it appeared from a vast distance or it appeared only through its effects or footsteps. The word *ruins* (أطلال) means the left signs after the original object had disappeared.

We will stop our analysis here despite the fact that all meaningful letters in Arabic and various different words were studied and analyzed according to our theory. The digital alphabet was tested with various one-, two-, three-, four- and five-letter words. We tested more than 150 words with positive and encouraging results. The logical and mathematical analysis of these words gave exactly the same abstract meaning. Sometimes it is quiet easy to figure out the word meaning from its abstract description, i.e., the word speaks by itself. Many times, it is not that straightforward and it requires a bit of imagination to link the abstract word with its everyday use.

## Conclusion and Future Remarks

We presented in this paper a new formal definition of Universal Grammar (UG). Instead of researching human brain, natural languages, genetics, history and psychology for a UG, we decided to take a different path and look for the basic logic behind a UG and a language, which is the communication and knowledge transfer. We developed a theory for communication and knowledge dissemination through language. This theory provides a rigorous mathematical definition of UG that complies well with many other grand theories in physics, astronomy, philosophy, mathematical logic and computer science. Thus, allowing any language, governed by this unique UG, to be fully analyzed, decoded and comprehended.

After formulating our UG and what we call the *Digital Alphabet*, we set out to look for a possibility of any natural language meeting our UG requirements. By a stroke of luck, our first trial using Arabic was positive so far. We got results that clearly show how Arabic words are constructed to give their precise meaning, giving us a lot of insight into how language, and communication in general, function. The presented work is a small part of an extensive research that has been carried out on this topic, sporadically, in the last three decades. More research have been done to understand how UG succeeds in delivering the right information, especially with regard to space and time. This resulted in extensive analyses of many grammatical features of Arabic language that touched directly with many open problems in semantics and pragmatics. Although, no result is conclusive, we think this new vision of UG and language is a gold mine of research that can be exploited in the upcoming years.